# Chapter 2

# Automatic video scene segmentation based on spatial-temporal clues and rhythm

Walid Mahdi and Liming Chen
*Département MI, Laboratoire ICTT, Ecole Centrale de Lyon, France*

Mohsen Ardebilian
*Laboratoire HEUDIASYC CNRS/UTC, Compiègne, France*

## 1. Introduction

With ever increasing computing power and data storage capacity, the potential for large digital video libraries is growing rapidly. These libraries will have thousands of hours of video, which will be made available, via the wide area networks, to users upon request [FAU 96, SMI 95].

However, the massive use of video for the moment is limited by its opaque characteristics. Indeed, a user who has to handle and retrieve sequentially needs too much time in order to find out segments of interest within a video. Therefore, providing an environment both convenient and efficient for video storing and retrieval, especially for content-based searching as this exists in traditional text-based database systems, has been the focus of recent and important efforts of a large research community [ARD 99, ARU 95, MAH 00].

The video coding scheme defined by MPEG-4 and the future MPEG-7 standards [MPE 99] offers several content-based functionalities, demanding a new organization of the video for better content description and understanding. Indeed, in order to enable content-based access of digital video libraries, the semantic and temporal structure embedded within a video must first be segmented, then indexed and searched with satisfactory recall and precision.

The majority of the existing automatic content-based segmentation of video consists of splitting the video into shots which are separated by cuts, and their detection is based on objective visual primitives such as color [SWA 91], image correlation, or 3-D hints [BER 89, ARD 00]. The shot represents the fundamental unit of manipulation of the video. So, by detecting the shot boundaries, it is possible to create indexes and to develop users' browsing tools enabling navigation and searching.

Unfortunately, the representation by shots of a video document does not describe its narrative and visual content well. Indeed, in such a case, the results of a user's query to a video library are a set of non-continuous time shots. They do not describe the narrative content of the video. Moreover, browsing thousands of shots contained in a video document (3225 shots in *October* of S.M. Eisenstein [AUM 94], 500 to 1000 shots in a normal film [AIG 94]), presents a great challenge for the user *during* his linear navigation and search for any particular video segment. Also the granularity of the information content is such that a one half-hour video may have ten semantically separate units, to each of which corresponds a scenario of the story board of the film.

Many approaches are proposed to improve search and navigation in the video medium. Their main idea consists in partitioning video data into small clips of meaningful data. A manual segmentation into 1600 video clips of a 45-hour library is considered by the Informedia Project [CHR 97]. Mills imposes a fixed segmentation into clips on the video data [MIL 92]. Ramin Zabih et al. describe in [RAM 95] an approach to the detection and classification of scene breaks in video sequences. Another a priori model has been proposed in [ZHA 94] for news broadcasts to extract certain specialized semantic elements of the video sequences. In our previous work, we proposed a method to detect strict alternate shots which are then combined into scenes [FAU 96].

In this paper, we propose a new automatic video scene segmentation method that explores two main video features; these are spatial-temporal relationship and rhythm of shots. The experimental evidence we obtained from a 80 minute-video showed that our prototype provides very high accuracy for video segmentation.

The rest of the paper is organized as follows. In Section 2, we briefly review related works in the literature, emphasizing strengths and drawbacks of each technique. In Section 3, we first define the problem by describing the video structure, we then present the general principle of our scene segmentation approach. Section 4 summarizes the 3-D hints-based algorithm used in our experimentation for video shot segmentation and parameterization. In Section 5, we introduce the clustering of video shots based on visual contents. Section 6 addresses the video scenes segmentation method based on spatial-temporal clues. A formulation of rhythm effect for video scenes segmentation is presented in Section 7. In Section 8, we detail our video scene segmentation method which couples the exploration of spatial-temporal relationships with the consideration of the rhythm effect. The experimental results that we have driven on a 80 minute-video selected from the movies *Dances With Wolves*, *Conte de Printemps* and *Un Indien dans la Ville* are then presented in Section 9. Further work is depicted in the last section.

## 2. Related work

While there is a lot of work reported in the literature focusing on shot boundary detection [AIG 95, ARD 00, CHA 94], there is, on the other hand, little work on the video scene segmentation. The first approach proposed the scene break detection on the basis of the image intensity signal analysis of the video stream. For instance, Otsuji and Tonomura [OTS 94] discuss a variety of measures based on the difference and change of intensity histograms of successive images. Nagasaka and Tanaka [AKI 94] present an algorithm using similar measures. Ramin, Miller and Mai propose a method in [RAM 95] based on the detection of intensity edges that are distant from edges in the previous frames. The limitation of this method appears clearly in the case where the video produces a rapid frequency of brightness changes which occurs for the total duration of the scene. In summary, the limit of these techniques is that they are only based on the spatial indication of the video stream and do not take into account the video temporal clues. Indeed, the global visual contents of successive images are not enough to differentiate different contexts. For instance, in the case of 'busy' scenes, the intensities may change substantially from frame to frame. This change often results from abrupt motion and leads to falsification of the scene break detection for these techniques based on similarity measures.

Jain Hampapur and Weymouth proposed another method [JAI 88], called *chromatic scaling*, which attempts to detect a variety of scene breaks based on an explicit model of the video production process. Their method implies the definition of a specific model for each type of video being analyzed. A particularly interesting approach is introduced by M. Yeung and Yeo [MIN 96, MIN 97]. For the first time they propose video shots clustering based on both visual similarities and temporal localities of the shots which they call *time-constrained clustering*. The goal of this method is to build a scene transition graph which provides the basis on which analysis can be performed to extract semantic story units. Unfortunately, the temporal locality criterion used in this method is difficult to generalize. Indeed, the choice of the temporal threshold, which imposes every scene to take place for a given time duration does not correspond to any video production rule. In our approach, we do not need such a temporal threshold. Instead, we use cinematic production rules to achieve scene break detection.

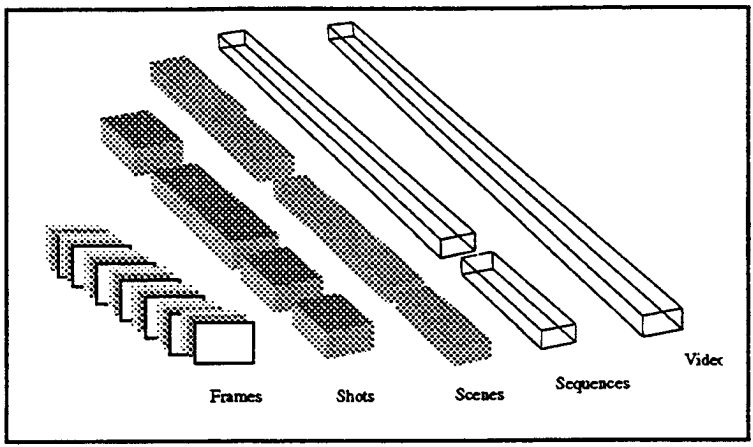

*Figure 1.*  Video structure

## 3. Video parsing: problem and principle

In this section, we first describe the semantic structure embedded within the video in order to clearly state the problem, then we introduce the principle of automatic scene segmentation within the framework based on spatial-temporal clues and rhythm.

### 3.1. Video structure

A video program such as motion pictures, TV movies, etc., has a story structure and organization. As illustrated in Figure 1, three levels define this syntactic and semantic story structure: narrative sequence, scene and camera shot. A camera shot is a set of *continuous frames representing a continuous action in time or space*. It represents the fundamental unit of production of video, reflecting a basic fragment of story units. A scene is a dramatic unit composed of a single or several shots. It usually takes place in a continuous time period, in the same setting, and involves the same characters. At a higher level, we have the narrative sequence, which is a dramatic unit composed of several scenes all linked together by their emotional and narrative momentum [AUM 94]. During the montage, which *refers to the editing of the film, the cutting and piecing together of an exposed film in a manner that best conveys the intent of the work*, two narrative sequences are linked together by an effect of transition such as dissolving or fading and two shots are linked by cuts. Consequently, this *temporal delimitation of sequences* criterion must be integrated in any model of video stream segmentation into semantic units.

    To convey parallel events in a scene, shots of the same person or same settings, taken from the same camera, at the same location, are repeated, alternated or interleaved by other shots with different contents. Most often, the similarities of contents are shown through similar visual characteristics of the composition of frames in the shots.

    Video is also characterized by a rhythm. In fact, certain editors produce rhythmical effects by using the duration of successive or alternative shots. Shots belonging to the same scene generally have the same or close rhythm. Moreover, shots which do not respect the current rhythm frequently carry out a scene break [AUM 94].

    For automatic video parsing purposes, we first introduce an intermediate level of *clusters*, resulting in a *scene*, then a *sequence* which provides narrative unity as follows.

**Definition 1 – Cluster**
*A cluster is a set of similar plans situated in a limited period of time.*

Similarity generally implies a similar set having the same spatial distribution. For example, let $\Gamma$ be the corpus of a film segmented into a set of shots $\Omega$. Consider 5 consecutive shots in $\Omega$: *p(k), p(k+1), p(k+2), p(k+3), p(k+4)*. If *p(k), p(k+2)* and *p(k+3)* describe such a similarity on set, actors, etc., while *p(k+1)* and *p(k+4)* describe another similarity, then two clusters *ϕ(i)={ p(k), p(k+2), p(k+3)}*, and *ϕ(i+1)={p(k+1), p(k+4)}* are formed.

Based on this definition, we can define more precisely the notions of scene and sequence.

**Definition 2 – Scene**
*A scene is a semantic unity formed by a set of successive shots temporarily continuous, sharing generally the same objects and/or describing the same subject.*

Let *ϕ(i)={ p(k), p(k+2), p(k+3)}* and *ϕ(i+1)={p(k+1), p(k+4)}* be two clusters sharing the same subject such as a telephone conversation and for which there exists no transition effect, then a scene *S(j)={p(k), p(k+1), p(k+2), p(k+3), p(k+4)}* is formed by the combination of *ϕ(i)*, and *ϕ(i+1)*.

Temporal continuance is defined by a non-interruption of the temporal projection of elements (shots) of this continuation (neither a plan miss nor an effect of transition exists between two consecutive plans). In other words, a scene is a semantic unity formed by a combination of additional clusters.

**Definition 3 – Sequence**
*A sequence is a narrative unity formed with one or several scenes.*

**3.2. Principle**

According to our approach, the video scene segmentation process consists of 3 steps: shot segmentation and parameterization, video shot clustering, and scene extraction. Figure 2 illustrates such a general process.

The shot segmentation decomposes the video into basic shots by detecting shot transition effect. Now there exists a lot of methods in the literature [ARD 00, RAM 95] for video shot segmentation. In this paper we suppose that shots and shot transition effects of a video are already segmented and are the input of our study. In our experience, we applied a 3-D hints-based method that we previously proposed [ARD 00]. A summary of this technique is presented in Section 4.

Video shots clustering aims at characterizing the spatial-temporal relationships between shots, and results in a set of clusters and a *temporal clusters graph (TCG)* capturing the temporal relationship between clusters.

Finally, we proceed to the scene segmentation. We first explore the *TCG* in order to get a rough scene segmentation. This result is then improved by the consideration of rhythm to obtain the final scene segmentation.

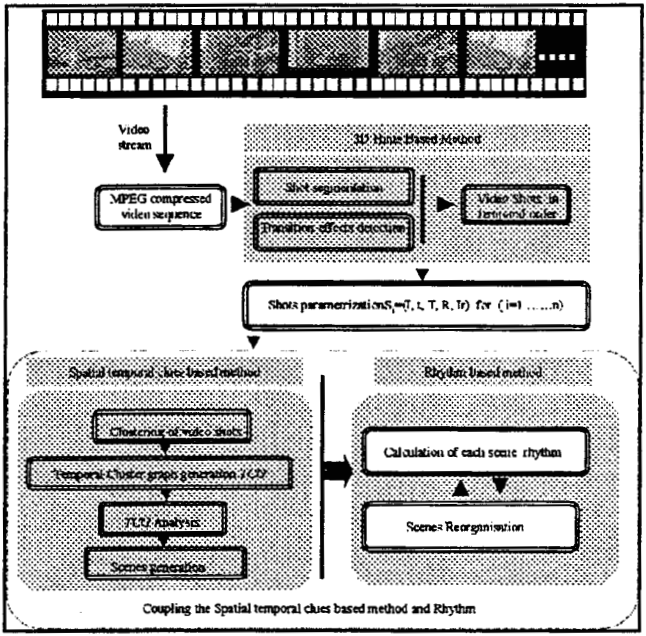

**Figure 2.**  *The block diagram of proposed approach for video scene segmentation*

## 4. Video shots segmentation and parameterization

The understanding of image content using 3-D clues has been the focus of many researchers working in the image-processing field [MAG 96, QUA 88, ZHA 94]. Not only can such knowledge contribute to information compression, eventually leading to video compression, but also to classify the type of the scenes with their 3-D structures [AUM 94, CHE 94]. In this section, we briefly describe our method of video shot splitting by the use of the *focus of expansion (FOE)* in 3-D scenes. According to our method, we first realize linear contour detection. Then by the principle of duality which exists in projective geometry, we localize the FOE within an image by a *Double Hough Transformation (DHT)*. Finally, with the help of FOE position change in the video stream, we split the video into shots. It should be noted that color information is not used in our technique, because it may alter the method robustness when sudden illumination or color changes occur in the video.

*Focus of expansion (FOE)* can be used to derive 3-D structure information such as parallelism, orthogonality of line features. Quantitative analysis of FOE gives information about motion as well. After double HT, we obtain for each input image a pattern of discrete points (possible FOE points). The positions of these points are used as indices of shot segmentation of the video sequence. In two-point patterns derived from two successive images, we have developed a matching algorithm for calculating the global match between two discrete point patterns. The main purpose is to find a maximum resemblance pattern. Here is the algorithm of the discrete point comparison:

1. Choose between two patterns the one with the lowest number (n) of points.
2. Superpose two patterns and do iteration on the points for i=1 to n.
3. Calculate the maximum resemblance of a pair of points in a fixed sized zone; if no corresponding point is found, assign a penalty value.
4. Go to step 2 to continue for all pairs of points.
5. Calculate the global resemblance of two patterns by summing all local values normalized by n.
6. By the help of the shot transition models take a decision if there is a shot transition in the video sequence or not and identify its type (cut, dissolve, etc).

This algorithm is very fast for a direct comparison between two patterns. We have also tested the algorithm by dynamic programming, and graph matching with its node representing the points and the links representing the relation among the points. However, to be efficient, the algorithm above must be carefully tuned to the global threshold value for shot cut detection, as with the majority of all other shot detection techniques.

## 5. Clustering of video shots

In order to organize the video data as semantic units, we first characterize the spatial-temporal relationship within the video by clustering similar shots. Recall that a cluster of shots gathers a set of similar shots and represents a fragment of semantic units. Thus two shots are clustered together if the distance of dissimilarity between them is less than a predefined threshold.

### 5.1. Similarity of images

Similarity measures based on visual variables such as color, image correlation, optic flow, 3-D images, etc., can distinguish different shots to a significant degree, even when operated on much more compressed images such as DC images. We adapt the histogram intersection method from Swain and Ballard [SWA 91] to measure the similarity of two images based on color histograms. A color histogram of an image is obtained by dividing a color spectrum (e.g. RGB) into discrete image colors (called *bins*) and counting the number of times each discrete color appears by traversing every pixel in the image. Given two images *Ii* and *Ij*, and their histograms *Hi* and *Hj*, each containing *n* bins, we denote by *Hi(k)* the value of *Hi* in the *k*th bin, the intersection of the histograms is defined as:

$$\sum_{k=1}^{n} \min\left(H_i(k), H_j(k)\right) \quad [1]$$

This gives an indication of the number of pixels that have similar colors in both images. It can be normalized to match a value between 0 and 1 by the following formula:

$$S(H_i, H_j) = \frac{\sum_{k=1}^{n} \min(H_i(k), H_j(k))}{\sum_{k=1}^{n} H_j(k)} \quad [2]$$

For our purpose, we use mismatch value, calculated from equation (3), to indicate dissimilarity between the two images instead of the match value defined above.

$$D(H_i, H_j) = 1 - S(H_i, H_j) \quad [3]$$

### 5.2. Temporal-delimitation for clustering of video shots

The clustering of video shots is aimed at grouping shots having similar content. For this purpose, we use the spatial-visual similarity combined with a temporal delimitation. Indeed, because of the temporal locality of the contents, if two similar visual shots occur far apart in the time line, they tend to represent different content or locale in different scenes. Each cluster of shots must be then in the same sequence to maintain semantic clustering. Generally, a sequence as described in [AUM 94], is linked to another one by an effect of transition (dissolve, fade-in or fade-out) to tell the spectator that a break in the narration of the history has occurred. Thus the process of clustering has only to be applied to each sequence separately. When all shots of a sequence are clustered a clustering of new sequence shots is brought out.

For the purpose of clustering, we assume that the shots and their parameters (*t*: time code, *Td*: time duration, *R*: representative image, *R*: effect of transition) of a video are already detected by means of the 3-D hints-based method. We also define *L(Si)* a function which returns the transition effect *R* used to link shot *Si* with the shot *Si+1*. Our strategy of clustering consists of initializing a new cluster φk to the first shot Si which is not yet classified, then it compares this one *(φk)* to each successive shot *Sj(j>i)* that is in the same sequence (*L(Si)*∉{dissolve, fade-in, fade-out}). If they are similar (*D(Si, Sj) < T* (threshold)), *Sj* is classified into φk. If not, the process of comparison is continued to the end of the sequence. When all shots of a sequence are clustered a new clustering of a sequence's shot is brought out.

### 5.3. Time-space graph generation

In order to organize the video data as a semantic unit, a new representation of the video data as a *time-space graph (TSG)* is then produced. The time-space graph is created as a new description model of the video stream. The TSG construction is based on shot boundaries detection and video shots clustering strategy as presented in the previous section. The vertical axis represents the spatial distribution of shots, the position of the cameramen or the clusters, with the horizontal representing the time line or the time code. Figure 3 illustrates a form of TSG, where τ is the time duration of an effect of transition between two sequences, $t_i$ is the time code of the shot $P_i$ and $\phi_i$ is a cluster that regroups some shots.

## 6. Video scenes extraction

Once we obtain the video clusters and characterize their temporal relationships, a new representation of the video stream, called *temporal-clusters graph (TCG)*, is built to carry out such temporal relationships between clusters. In contrast with the scene transition graph generated by M. Yeung and Yeo [MIN 96], the *TCG* generation is based on temporal presentation of multimedia programs. Video scenes are then extracted by analysis of temporal relationships on the TCG. For temporal presentation of multimedia programs, *Timing Petri-net graphs* [BUC 93] or *Allen temporal operators* [ALL 83] may be used. Timing petri-net is often used to obtain the expression of the synchronization of several multimedia objects such as audio stream and video stream within a video program. But the temporal modeling in such a graph may be complex for a long video program. In our work, as we only focus on the video stream, Allen operators are preferred because they offer the same expression power with greater simplicity.

### 6.1. Temporal-clusters graph generation

Based on TSG, temporal relationships between clusters are defined and captured by a temporal cluster graph TCG. To describe such temporal relationships between clusters, some new annotations are introduced for primitive temporal relations defined by Allen and described in [ALL 83]. In such a *TCG*, a node is associated with each cluster, and an edge is drawn for each temporal relationship between two clusters. Figure 4 illustrates an example of TCG. This representation defines a video as a *directed acyclic graph* [LAY 96] where edges are time constraints on the duration as defined in Table 1. These time constraints give the allowable duration of intervals which is specified by the parameters of temporal relation. We also make use of two abstract time-nodes, *Begin* and *End*, corresponding to the time of the beginning and the end of a given video program. The *Begin* node is connected to all time-nodes which do not have incoming edges by means of special edges labeled with a particular delay. Similarly, all the time nodes which have no outgoing edges are connected to the *End* time node. This DAG representation of a video is used as the basis for the time constraints analysis. Figure 5 shows an example of a DAG corresponding to our previous example presented in Figure 4.

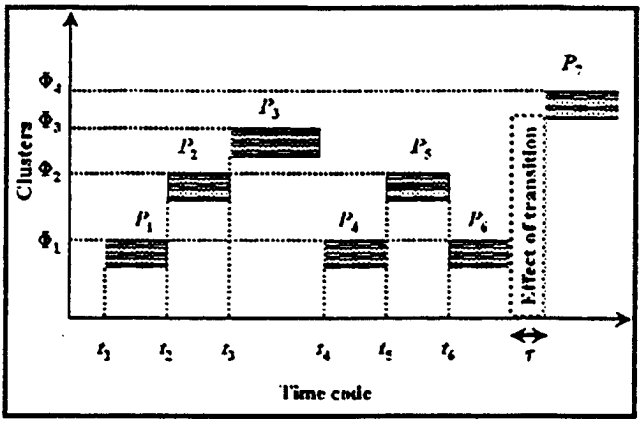

*Figure 3. Description of video data as TSG graph*

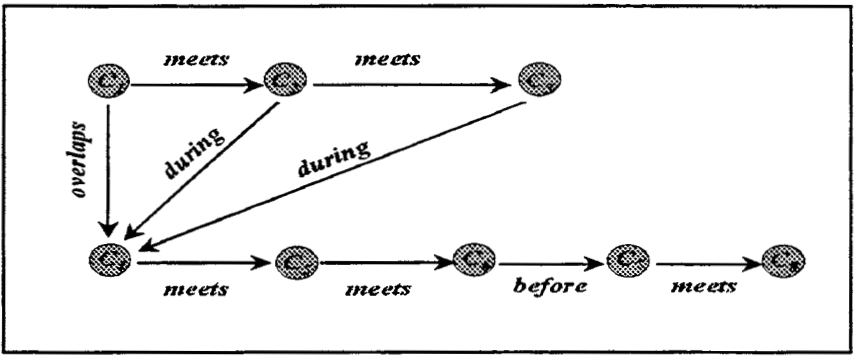

***Figure 4.*** *An example of TCG graph*

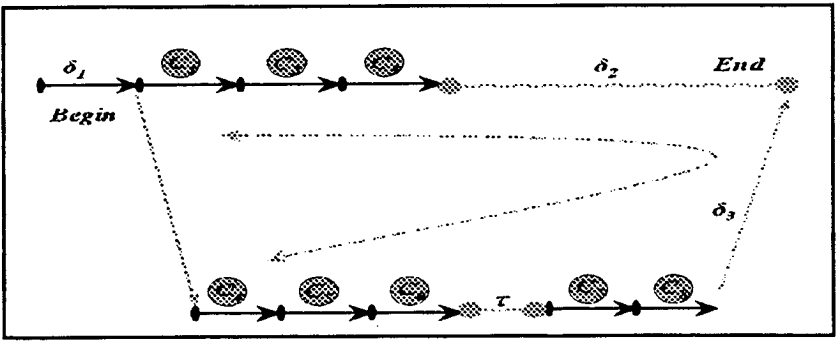

***Figure 5.*** *Temporal Directed Acyclic Graph generated from the TCG*

*Table 1.* Temporal relation's constraints

| Temporal relations | Temporal constraints |
|---|---|
| $\Phi_1$ Meets $\Phi_2$ | $P_n^1.t + P_n^1.Td = P_1^2.t - 1$ |
| $\Phi_1$ Before $(\tau)$ $\Phi_2$ | $P_n^1.t + P_n^1.Td + \tau = P_1^2.t - 1$ |
| $\Phi_2$ During $(P_i^2.t,...,P_i^2.t,...P_{i+1}.t)$ $\Phi_1$ | $P_n^1.t + P_1^1.Td \leq P_i^2.t - 1$ && $P_i^1.t + P_i^1.Td \geq P_n^2.t - 1$ |
| $\Phi_1$ Overlaps $(P_1^2.t,...,P_i^2.t,...P_{i+1}.t)$ $\Phi_2$ | $P_1^1.t + P_1^1.Td \leq P_1^2.t - 1$ && $P_n^1.t + P_n^1.Td \geq P_i^2.t - 1$ && $P_n^1.t + P_n^1.Td \leq P_{i+1}^2.t$ |

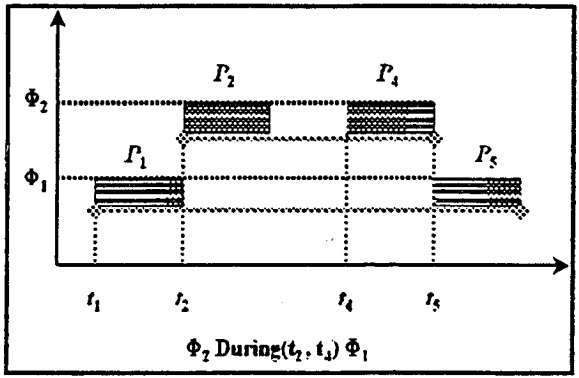

*Figure 6.* During relationship generated from the TSG graph

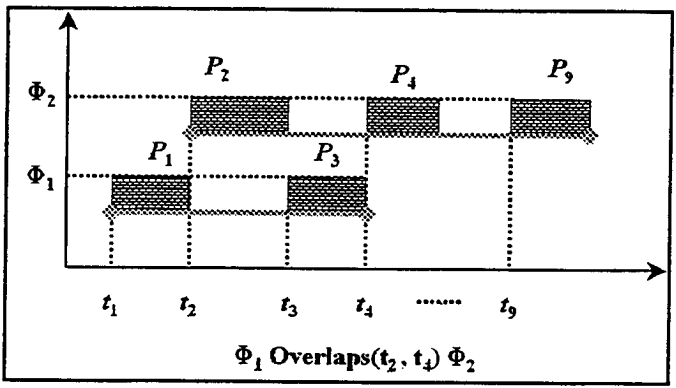

*Figure 7.* Overlaps relationship generated from the TSG graph

Given an arbitrary DAG representation extracted from the temporal cluster graph, the following properties can be stated:

- Certain routes are not allowed (Acyclic Graph) because *instants coming from the past can not be equal to instants in the future* and vice versa.
- Cycles are always composed of two parallel chains (*DAG property*) [LAY 96], whose total duration must be compatible. This property ensures that the two paths can be guaranteed at run-time, as illustrated in Figure 5.

Currently, we used four types of relations between clusters: *Meets, Before, During* and *Overlaps*. The first two describe a sequential relationship between two clusters. Therefore, they have only one parameter that represents the delay $\tau$ of succession between them. The *Meets* relation is generated between two clusters that are included in the same sequence, and no delay separates these two clusters. The *Before* relation describes a transition of sequences. Each member of this relation is included in a different sequence, then in a different semantic unit. The delay parameter represents the time duration of the effect of transition that links the two sequences together. *During* and *Overlaps* are used when two clusters intersect in time, as illustrated in Figure 6 and Figure 7. They indicate that their members are included in the same semantic unit. The time code of all shots of the left cluster of a *During* relation defines its parameters. For an *Overlaps* relation, its parameters are defined by the time code of all shots of its right-hand member in intersection with its left-hand member. In addition, the last parameter of this relation is the time code of the first shot of the right-hand member which is not in intersection with the left-hand member as illustrated in Figure 7. The succession and the intersection between two members of a temporal relation are determined by the space-time distribution of their shots. As a consequence of the previous definitions, these annotated temporal relations define a set of temporal constraints which must be satisfied. Table 1 summarizes all of these temporal constraints associated with the temporal relations. In the table, $t$ represents the time code of a shot, $Td$ the time duration of a shot and $\tau$ the time duration of the effect of montage that links one shot to its successor (if *cut* then $\tau = 0$ otherwise $\tau \neq 0$).

For the extraction of scene units, we first analyze the TCG according to the semantic of the temporal relationship between cluster nodes, leading to a first scene segmentation result. This result is then improved by the consideration of the rhythm effect to obtain the final scene segmentation.

### 6.2. Temporal-clusters graph analysis

Extracting semantic units of a video document consists of analyzing the four types of temporal relationships captured by a TCG: *Before, Meets, Overlaps* and *During*. Each one corresponds to some rule in the video production. We first describe these semantic correspondences:

- *Meets*: this separates two clusters which belong to two successive scenes.

These scenes belong to the same sequence. For instance, consider two shots *a* and *b* that belong to two different and successive scenes. Suppose that according to the shots clustering method as presented in the previous section, these two shots are classified in two different clusters. However, if the effect of montage that links shot *a* to shot *b* has a time duration equal to 0, a *Meet* temporal relationship is generated in order to separate two successive scenes.

– *Before*: this separates two clusters that belong to two scenes of two successive sequences. This relationship occurs between two successive clusters when a montage feature is used, such as fade, dissolve, or using a time duration different from zero.

– *Overlaps* and *During*: these describe a change in the temporal distribution of shots belonging to the same scenes. In other words, this type of relationship corresponds to the case where movie editors interleave, by means of a montage, several shots in order to describe a scene which takes place with a different setting but having a common subject. A classic example of using an *Overlaps* relationship is the telephone conversation.

We turn to the video segmentation process based on spatial-temporal relationships. Following from the previous analysis, we can see that the *Before* relationship divides a video into several sequences, and is used to separate a TCG into several independent subgraphs, called *sequence subgraphs*.

Figure 4 illustrates a TCG which is split into two subgraphs by the *Before* relationship between clusters 6 and 7. Consequently, the semantic scene units must be separately searched in each *sequence subgraph* of TCG. For this purpose, the *Meets* relationship is used to divide each sequence subgraph into scene units, $\Gamma$, which are in their turn subgraphs of the sequence subgraph. Each subgraph $\Gamma$ contains at least one shot. A series of successive shots obtained by the combination of all nodes of subgraph $\Gamma$ describes the boundaries and the content of a semantic scene unit. The application of this process leads to a first scene segmentation result. We experimented with this process on a 45 minute-video from the movie *Dances With Wolves*, 15 minutes from the movie *Conte de Printemps*, and 20 minutes from the movie *Un Indien dans la Ville*. The experimental results are presented in Tables 5, 6, 7 and 8 of Section 9. As we can see from these tables, this process, which reaches a high success segmentation rate, has also the drawback of producing independent scenes formed only by one shot. This situation corresponds to a distortion of the real semantic of the video. This consideration leads us to apply an additional tool based on the rhythm which allows capture of the complete semantic embedded within a video.

## 7. The use of rhythm

Rhythm is a temporal effect that editors produce by using the duration of successive shots, leading to particular visual effects for each scene. For instance, a sunset scene in a video has a different rhythm as compared to that of an air-raid scene. Furthermore, the rhythm change between scenes is often dependent on the movement within scenes. Our previous example illustrates this fact. Sun in a sunset scene moves more slowly than two aircraft of an air-raid scene.

To explore the rhythm effect within a video stream, we first propose a statistical measure to detect the change or maintenance of the rhythm. We then discuss how such a measure can be used for our scene segmentation purpose.

**7.1. Statistical characterization of rhythm**

Experimental results show that shot duration in video can not be determined in advance and variation of shot duration follows a random distribution. Thus we may approximate a set of shot duration variations as a set of statistical variables. In order to describe the distribution of a statistical variable, we rely on a theoretical distribution which characterizes such a statistical variable according to a law of parametric probability.

The law of parametric probability which any statistical variable obeys is generally deduced from theoretical considerations and hypothesis. In our case, with the aim to study the distribution law of the variation of the duration of different shots within a same scene, we represent the whole set of these variation points on the plan $(v, f)$, where $v$ stands for the variation of the duration of shots, whilst $f$ stands for the frequency of each variation. Figure 8 shows the histograms which correspond to different scenes extracted from several films. We can notice from the figure that the different histogram curves match a well-known distribution rule which has the shape of a bell. This rule is called in statistics the normal law of distribution [SAB 93]. However, stating from a sample of videos that the variation of duration of shots in every film follows the normal law of distribution may lead to some error. Consequently, we make our assessment in a so-called *safe interval* which guaranties to a certain degree, that a point can be applied to a population of a statistical variable. In our case, we mean by population of a statistical variable a succession of video shots, and by *safe interval* the interval which statistically characterizes the two ends of a semantic scene.

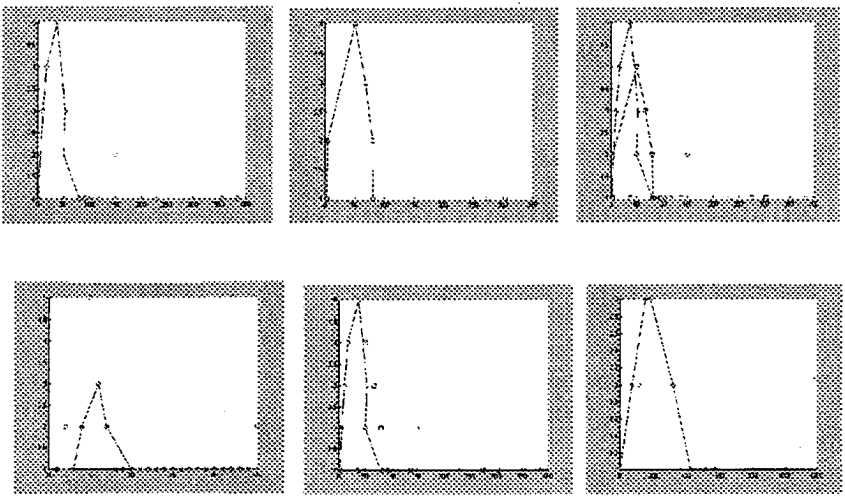

*Figure 8.* Shot duration variation distribution; **a, b** and **c** three scenes from the film Dances with Wolves, **c** and **d** two scenes from the film Conte de Printemps, **e** and **f** two scenes from the sitcoms provided by INA

**7.2 Scene segmentation by the use of rhythm**

Assume there exist N shots in a video $\varphi = \{P_1, P_2, \ldots, P_i, P_{i+1}, \ldots P_N\}$. Every shot $P_i$ ($1 \leq i \leq N$) is characterized by its time duration $Td_i$ and its time code (starting time) $t_i$. Assume also that $VTP_{i,i+1}$ is the duration's variation between shots $P_i$ and $P_{i+1}$, that is $VTP_{i,i+1} = |P_i - P_{i+1}|$. Thus, to the set of shots $\varphi$, we associate a set of duration variations between each couple of successive shots $P_i$ and $P_{i+1}$ (with $P_i, P_{i+1} \in \varphi$).

Let $GP = \{P_1, P_2, \ldots, P_n\}$ be a set formed by the first $n$ ($n>2$) and successive shots. Let $GVTP = \{VTP_{1,2} \ldots VTP_{n-1,n}\}$. We define $F(GP)$ and $L(GP)$ as two functions which respectively give the first and the last shot belonging to $GP$. A new shot $P_i$ ($n+1 \leq i \leq N$) is considered to have the same rhythm as compared to shots in $GP$ and is added to $GP$ if it satisfies the two following conditions:

(a) The temporal continuity condition:
The insertion of the shot $P_i$ to $GP$ must ensure a temporal-continuity between shots $P_j$ ($P_j \in GP$), that is : $P_i.t + 1 = PF(GP).t$ or $P_i.t = PL(GP).t + 1$. This condition simply states that shot Pi must be the one which precedes the first shot or follows the last shot of $GP$.

(b) The aggregation condition based on the rhythm:

$$|VTP_{F(GP), i} - VTPM| \leq \alpha * \delta \quad \text{or} \quad |VTP_{L(GP), i} - VTPM| \leq \alpha * \delta$$

where:

– *VTPM* is the average variation of the *GVTP* set, and is calculated by equation 4.

$$VTPM = \frac{\sum_{i=1}^{n-1} VTP_{i,i+1}}{n} \qquad [4]$$

– $\delta$ is the standard deviation of the set *GVTP*, and is calculated by equation 5.

$$\delta = \sqrt{\frac{\sum_{i}^{n-1}(VTP_{i,i+1} - VTPM)^2}{n}} \qquad [5]$$

– $\alpha$ is a coefficient that determines the *safe interval* $I = [VTPM - \alpha\delta, VTPM + \alpha\delta]$.

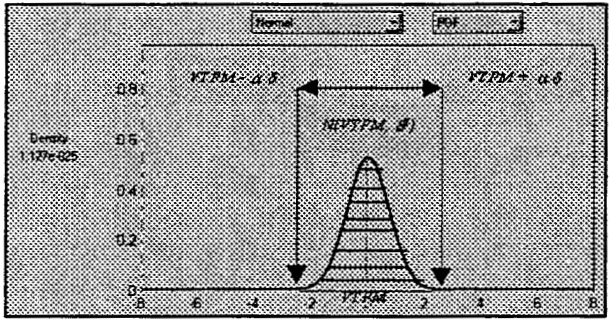

**Figure 9.** *The curve of a Normal Law density function*

Suppose that the set *GVTP* of rhythm changes is characterized by the curve of a normal low-density function as illustrated in Figure 9. If α is fixed to be value 2.5, and if the addition of a shot $P_i$, to the set *GVTP* leads to a new value of *VTPM* contained in *I*, then according to statistics theory, shot $P_i$ will have a probability of 98% of being an element of the scene *GP*. When one of the two previous conditions is not satisfied, a new set *GP* is initialized with *n* (*n*>2) successive shots, and the process of shot grouping is tried again with the rest of shots still untreated. Generally, the value n is chosen according to the synoptic character that the segmentation process has to respect. This value generally depends on the type of video program. For instance, a *TV* news frequently comes back to the anchor person after a report which may be only composed of two shots. In this case, it would be necessary to choose for *n* a value corresponding to the minimum number of shots that a scene may have.

In our experience we have fixed *n* = 3 because of the synoptic character of the dramatic films on which we have conducted our experimentation. However, we rarely base our scene segmentation process just on the rhythm rule as we first use spatial-temporal clues which lead to an initial semantic shot aggregation, and thus the set of *GP* already formed. The following section explains how we couple the rhythm with spatial-temporal clues to more precisely capture temporal structure within a scene and between scenes in a video, to give the final scene segmentation.

## 8. Coupling the use of rhythm with spatial-temporal clues-based segmentation

Our segmentation process based on spatial-temporal clues [MAH 98], when applied to several samples of videos may lead to the creation of a so-called shot scene which is a scene formed by one shot. This is often a distortion of the semantic. Rhythm is used here to tackle this problem. Indeed, each shot scene is studied on the basis of rhythm in order to possibly be associated with the scene which immediately precedes or follows it. This coupling-based segmentation is achieved as follows:

- First, we apply the spatial-temporal clues-based segmentation to the video stream in order to obtain an initial result of scene segmentation.
- Let $\Gamma = \{S_1, S_2 \ldots S_N\}$ be the set of scenes containing more than one shot.
- Let $\chi = \{P_i, \ldots P_j\}$ be the set of one shot scenes. Thus, $\forall P_i \in \chi$, and we have $\neg \exists S_i \in \Gamma$ such $P_i \in S_i$.

- Now each $S_i$ in $\Gamma$ is considered as a set of shots $GP_i$ as already defined in the previous section. Notice that we do not have to choose a particular value for $n$ and obtain a $GP$ set already computed. We depict the informal algorithm as follows:

**Input**: Scenes $S_1 \ldots S_n$ already obtained by the spatial-temporal clues-based segmentation.
**Step 0**: Initially, there are $n$ set of scenes $GP_1 \leftarrow S_1; \ldots; GP_n \leftarrow S_n$;
**Step 1**: Let $\chi_r$ be the set of one shot scenes; $\chi_i \leftarrow \{\}$;
//$\chi_i$ contains the previous state of $\chi_i$//
**Step 2**:
**If** ($\chi_r \neq \{\}$) //there is any one shot scene ?//
   **2.0.** $\chi_i \leftarrow \chi_r$;
   **2.1.** Select scene $GP_i$ formed by one shot from $\chi_r$;
   **2.2.** Look in $\Gamma$ for all scenes $GP_k$ which immediately precede or follow the one shot scene $GP_i$, i.e. $P_i.t = P_{F(GPk)}.t - l$ or $GPi.t = P_{\mu(GPk)}.t + l$;
   Else// there is no one shot scene//
   **2.3.** Go to **step 5**;
  Endif;
**Step 3:**
**For** each $GP_k$ which has been chosen at **step 2.2**
   compute the set $GVTP_k$;
   compute $VTPM_k$;
   compute $\delta_k$;
   //check the aggregation condition//
   **If** ($|VTP_{L(GPk)}._i - VTPM_k| < \alpha * \delta_k$) or ($|VTP_{F(GPk)}.i - VTPM_k| < \alpha * \delta_k$)
     **If** ($GP_k$ corresponds to the scene that best matches the rhythm)
     $GP_k = GP_k + \{GP_i\}$;
     // the one shot scene is removed from the set $\chi_r$//
     $\chi_r = \chi_r - \{GP_i\}$;
     **Endif;**
   **Endif;**
**End;**
**Step 4**:
**If** ($\chi_r \neq (\chi_i)$ // is there any aggregation ?// go to **step 2**; **Endif**

Automatic video scene segmentation    35**Step 5**: Stop.
**Output**: scenes $S_1 \ldots S_p$ with $p \leq n$

Once the above algorithm is applied, there may be some groups of successive shots which remain and do not belong to any segment. They are then handled as follows:

– In case a group of successive shot scenes is formed by more than two shots *(n>2)*, the rhythm-based scene segmentation technique that we have introduced in Section 7 is applied to obtain a synoptic segmentation.

– Otherwise each shot scene is considered as a scene which is often used in order to emphasize a jump in the narration or a jump in the film time.

## 9. Experimental results

In order to show the efficiency of our coupling model for the semantic scene segmentation purpose, we have implemented a prototype in *Matlab* and *Visual C++* using the standard collections *STL* on a *Pentium-II* based machine. We have driven experiences on segments extracted from three movies: 15 minutes of *Conte de Printemps*, 20 minutes of *Un Indien dans la Ville*, and 45 minutes of *Dances with Wolves* with Kevin Costner. They have been chosen for the variety of style. All the sequences are compressed in MPEG-1. In our experiences, all MPEG sequences were first decoded and segmented into parameterized shots by the 3-D hints-based method [ARD 00]. These parameterized shots are then used as an input to our study. Each shot is represented by a key frame at a spatial resolution of 288 × 352 with each pixel coded by 32 bits (16M colors). A summary of the shot segmentation results is shown in Table 2. We can see in Table 2 that the first segment of the movie *Dances with Wolves* contains two dissolves detected at frame 70 and frame 98, which can be used as temporal delimitation for the scene segmentation. For the other segments, there is no transition effect, thus in this case only the rhythm is used for the temporal delimitation. Table 3 provides the scene segmentation results by three methods: spatial-temporal clues-based method, coupling method and expert-based manual segmentation. Column 4 in this table shows the number of clusters obtained by the clustering process. Recall that our clustering process is based on histogram color comparison and uses the measure of dissimilarity with the threshold $T = 0.1$. Figure 10 shows the result of clustering

**Table 2.** *Results of 3-D hints-based method for shot segmentation and parameterization*

| Type | Sequence Name | # Shots | # Dissolves |
|---|---|---|---|
| Movie segment | Conte de printemps | 67 | 0 |
| Movie segment | Un indien dans la ville | 85 | 0 |
| Movie segment (1) | Dances with wolves | 130 | 2 |
| Movie segment (2) | Dances with wolves | 150 | 0 |

*Table 3.* Results of sample test sequences

| Sequence Name | # Shots | # Frames | # Clusters | # Story units(1) | # Story units(2) | # Story units(3) |
|---|---|---|---|---|---|---|
| Conte de printemps | 67 | 13820 | 46 | 25 | 5 | 4 |
| Un indien dans la ville | 85 | 24120 | 63 | 32 | 6 | 5 |
| Dances with wolves (1) | 130 | 34020 | 69 | 23 | 10 | 8 |
| Dances with wolves (2) | 150 | 37500 | 99 | 5 | 4 | 3 |

(1): Story units segmented by the spatial-temporal clues-based methods.
(2): Story units obtained by the spatial-temporal clues-based methods coupled with rhythm
(3): Story units identified by an expert.

*Table 4.* Content of clusters

| Cluster | Frame numbers |
|---|---|
| C16 | 16 |
| C17 | 17 |
| C18 | 18 |
| C19 | 19, 24 |
| C20 | 20, 23 |
| C21 | 21, 22 |

after 15-minutes in *Conte de Printemps*. Table 4 shows the content of some clusters which result from the clustering process shown in Table 3.

These are used to generate the TCG, illustrated by Figure 11, which characterize the temporal relationships among clusters of the video. Story units (scenes) can be extracted by analyzing these temporal relationships as explained in Section 6.2. In Figure 11, we have a part of the TCG on the 15 minutes *Conte de Printemps* which describes the temporal relationships among clusters *16, 17, 18, 19, 20, 21* and *22* highlighted in Figure 10. Each edge in the graph in Figure 11 represents a temporal relationship between two clusters. Generally there may be four types of temporal relationship which are *Overlaps, During, Before* or *Meets*. We can observe in Figure 11 two temporal relationship types, *During* and *Meet*, describing the following relations:

1. C16 Meets C17
2. C17 Meets C18
3. C18 Meets C19
4. C19 Meets C22
5. C20 During (19, 22) C19
6. C21 During (20, 21) C19
7. C21 During (20, 21) C20

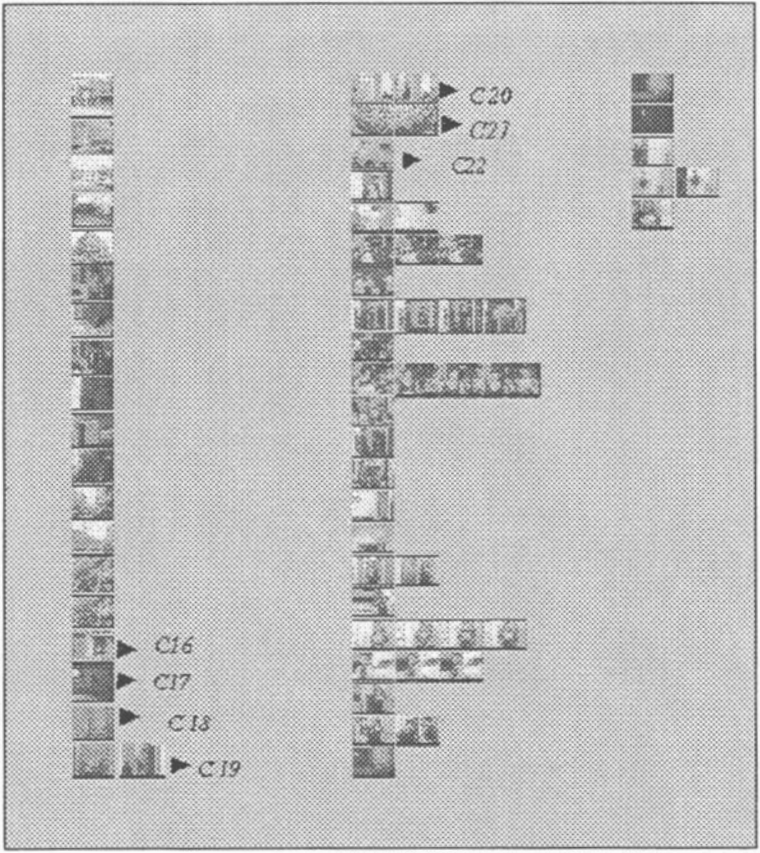

*Figure 10.* Clusters obtained after the clustering process on 15-minutes Conte de Printemps

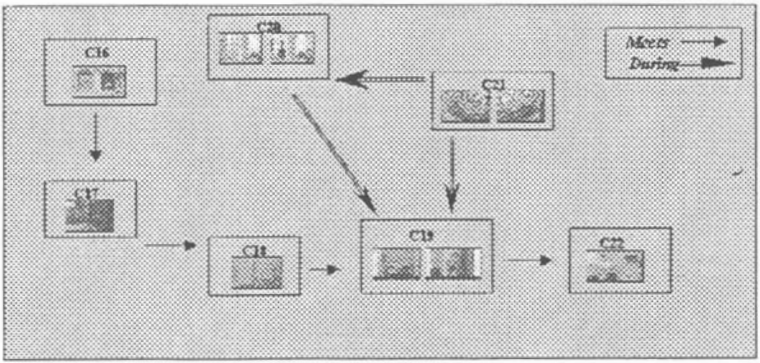

*Figure 11.* A sample of the TCG built on the 15-minutes Conte de Printemps

**38**   Video dataAs described in Section 6.2, the *During* relationship characterizes a change in the temporal distribution of shots belonging to the same scene. Therefore, clusters *C19, C20* and *C21* are merged into one scene. On the other hand, if we only rely on spatial-temporal relationships for the scene segmentation, we know that the *Meets* relationship occurs when two successive clusters belong to two different successive scenes. Thus clusters *C16, C17, C18,* and *C22* would each generate a scene. However, when looking into the video, an expert-based manual segmentation would gather *C16, C17* and *C18* into one scene with the one formed by *C19, C20* and *C21* because of their intrinsic semantic showing that one lady goes back to her home. Similarly for cluster *C22*. Fortunately, as illustrated in Figure 12, the analysis of rhythm gives other information which eventually leads to merger on one hand *C22*, and on the other hand clusters *C16, C17* and *C18* with the scene formed by clusters *C19, C20* and *C21*. As we can see in the figure, cluster *C22* is first merged into the scene (*C19, C20, C21*), then *C18, C17* and finally *C16*. Table 3 summarizes the scene segmentation results using three different methods. While column #7 gives the number of scenes resulting from an expert-based manual segmentation, columns #5 and #6 respectively give the results using the spatial-temporal clue-based segmentation method and the coupling segmentation-based method where we have fixed $T = 0.1$ and $\alpha = 2.25$. Tables 5, 6, 7 and 8 detail the scene segmentation results on the three movies, giving the content of each segmented scene by the three methods. As we can see in the tables, while the spatial-temporal relationship based segmentation method provides good results, it also produces many one-shot scenes. The use of the rhythm removes this drawback, leading to fully successful scene segmentation as compared to the expert manual segmentation.

***Table 5.*** *Experimental results on the 20-minutes Un Indien dans la Ville*

| Spatial temporal clue based segmentation method | | | Coupling method | | | Expert results | | |
|---|---|---|---|---|---|---|---|---|
| N° Scene | Begin | End | N° Scene | Begin | End | N° Scene | Begin | End |
| 1 | Shot 19 | Shot 24 | 1 | Shot 1 | Shot 14 | 1 | Shot 1 | Shot 14 |
| 2 | Shot 29 | Shot 44 | 2 | Shot 15 | Shot 25 | 2 | Shot 15 | Shot 26 |
| 3 | Shot 45 | Shot 66 | 3 | Shot 27 | Shot 44 | 3 | Shot 27 | Shot 44 |
|  |  |  | 4 | Shot 45 | Shot 66 | 4 | Shot 45 | Shot 66 |
| Remaining Shots considered as scenes | | | Remaining Shots considered as scenes | | | Remaining Shots considered as scenes | | |
| 1, 2, 3, 4,...18, 25, 26, 27, 28 | | | 26 | | | ------------------------ | | |

***Table 6.*** *Experimental results on the 15-minutes Conte de Printemps*

| Spatial temporal clue based segmentation method | | | Coupling method | | | Expert results | | |
|---|---|---|---|---|---|---|---|---|
| N° Scene | Begin | End | N° Scene | Begin | End | N° Scene | Begin | End |
| 1 | Shot 6 | Shot 12 | 1 | Shot 1 | Shot 5 | 1 | Shot 1 | Shot 5 |
| 2 | Shot 24 | Shot 56 | 2 | Shot 6 | Shot 16 | 2 | Shot 6 | Shot 17 |
| 3 | Shot 57 | Shot 65 | 3 | Shot 18 | Shot 56 | 3 | Shot 18 | Shot 56 |
| 4 | Shot 66 | Shot 73 | 4 | Shot 57 | Shot 65 | 4 | Shot 57 | Shot 65 |
|  |  |  | 5 | Shot 66 | Shot 85 | 5 | Shot 66 | Shot 85 |
| Remaining Shots considered as scenes | | | Remaining Shots considered as scenes | | | Remaining Shots considered as scenes | | |
| 1, 2, 3, ... 5, 13, 14,...,23, 74, 75, ...,85 | | | 17 | | | ------------------------ | | |

**Table 7.** Experimental results on the 20-minutes Dances with Wolves (CD-1)

| Spatial temporal clue based segmentation method | | | Coupling method | | | Expert results | | |
|---|---|---|---|---|---|---|---|---|
| N° Scene | Begin | End | N° Scene | Begin | End | N° Scene | Begin | End |
| 1 | Shot 1 | Shot 29 | 1 | Shot 1 | Shot 29 | 1 | Shot 1 | Shot 29 |
| 2 | Shot 33 | Shot 47 | 2 | Shot 31 | Shot 51 | 2 | Shot 31 | Shot 50 |
| 3 | Shot 53 | Shot 66 | 3 | Shot 53 | Shot 66 | 3 | Shot 51 | Shot 65 |
| 4 | Shot 71 | Shot 97 | 4 | Shot 67 | Shot 70 | 4 | Shot 66 | Shot 70 |
| 5 | Shot 101 | Shot 116 | 5 | Shot 71 | Shot 98 | 5 | Shot 71 | Shot 98 |
| 6 | Shot 117 | Shot 128 | 6 | Shot 100 | Shot 116 | 6 | Shot 99 | Shot 116 |
| | | | 7 | Shot 117 | Shot 130 | 7 | Shot 117 | Shot 130 |
| Remaining Shots considered as scenes | | | Remaining Shots considered as scenes | | | Remaining Shots considered as scenes | | |
| 30, 31, 32, 48, 49, 50, 51, 52, 67, 68, 69, 70, 98, 99, 100, 129, 130 | | | 30, 52 and 99 | | | 30 | | |

**Table 8.** Experimental results on the 25-minutes Dances with Wolves (CD-2)

| Spatial temporal clue based segmentation method | | | Coupling method | | | Expert results | | |
|---|---|---|---|---|---|---|---|---|
| N° Scene | Begin | End | N° Scene | Begin | End | N° Scene | Begin | End |
| 1 | Shot 2 | Shot 25 | 1 | Shot 2 | Shot 25 | 1 | Shot 1 | Shot 25 |
| 2 | Shot 26 | Shot 63 | 2 | Shot 26 | Shot 64 | 2 | Shot 26 | Shot 64 |
| 3 | Shot 65 | Shot 150 | 3 | Shot 65 | Shot 150 | 3 | Shot 65 | Shot 150 |
| Remaining Shots considered as scenes | | | Remaining Shots considered as scenes | | | Remaining Shots considered as scenes | | |
| 1 and 64 | | | 1 | | | ----------------- | | |

## 10. Conclusions and future work

We propose a new scene segmentation method, which couples the exploration of spatial-temporal relationships with the consideration of video stream rhythm. Our method, applied to a video, automatically decomposes the video into a hierarchy of scenes, clusters of similar shots and shots at the lowest level. Experience on a 80-minute video stream extracted from the movies *Dances with Wolves*, *Conte de Printemps* and *Un Indien dans la Ville* shows our method produces a very high scene segmentation success rate as compared with expert manual segmentation.

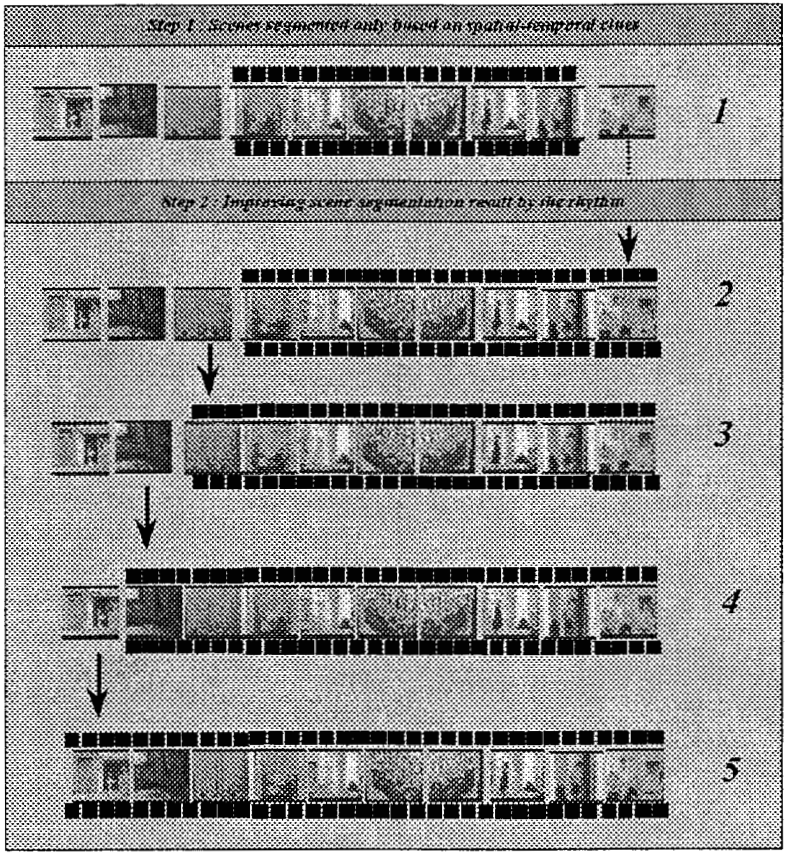

*Figure 12.* *Use of rhythm for rectifying the drawbacks of one shot scenes produced by the spatial-temporal clues-based segmentation*

Our future work includes the exploration of other spatial clues, for 2-D and 3-D impressions, for a better shot clustering, and definition of techniques for further video macro-segmentation based on sequence segmentation.

Recently, we have worked out two new techniques for the segmentation of other significant clues, that is image classification into *inside/outside* and *day/night* images, and the localization of texts embedded within a video program. Based on these techniques, we have been extending our scene segmentation work on other video materials such as commercials and TV news.